\documentclass[]{imag-ms-template}
\usepackage{amsmath}
\usepackage{todonotes}
\usepackage{graphicx}
\usepackage{caption}
\usepackage{subcaption}
\usepackage[normalem]{ulem}
\usepackage{booktabs,multirow}
\usepackage[percent]{overpic}

\DeclareMathOperator*{\argmin}{arg\,min}

\title{EEG-to-fMRI synthesis of task-evoked and spontaneous brain activity: addressing issues of statistical significance and generalizability}

\author{Neil Mehta,$^{1,2}$ Inês Gonçalves,$^{1}$ Alberto Montagna$^{1}$, Mathis Fleury$^{1}$, \\ Gustavo Caetano$^{1}$, Inês Esteves$^{1}$, Athanasios Vourvopoulos$^{1}$, \\ Pulkit Grover,$^{2}$ Patr{\'\i}cia Figueiredo$^{1}$ \\
{\small $^{1}$Institute for Systems and Robotics – Lisboa and Department of Bioengineering,} \\
{\small Instituto Superior Técnico, Universidade de Lisboa, Portugal}\\
{\small $^{2}$Electrical and Computer Engineering Department,} \\ 
{\small Carnegie Mellon University}
}

\begin{document} 

\maketitle 


\begin{abstract}
A growing interest has developed in the problem of training models of EEG features to predict brain activity measured using fMRI, i.e. the problem of EEG-to-fMRI synthesis. Despite some reported success, the statistical significance and generalizability of EEG-to-fMRI predictions remains to be fully demonstrated. Here, we investigate the predictive power of EEG for both task-evoked and spontaneous activity of the somatomotor network measured by fMRI, based on data collected from healthy subjects in two different sessions. We trained subject-specific distributed-lag linear models of time-varying, multi-channel EEG spectral power using Sparse Group LASSO regularization, and we showed that learned models outperformed conventional EEG somatomotor rhythm predictors as well as massive univariate correlation models. Furthermore, we showed that learned models were statistically significantly better than appropriate null models in most subjects and conditions, although less frequently for spontaneous compared to task-evoked activity. Critically, predictions improved significantly when training and testing on data acquired in the same session relative to across sessions, highlighting the importance of temporally separating the collection of train and test data to avoid data leakage and optimistic bias in model generalization. In sum, while we demonstrate that EEG models can provide fMRI predictions with statistical significance, we also show that predictive power is impaired for spontaneous fluctuations in brain activity and for models trained on data acquired in a different session. Our findings highlight the need to explicitly consider these often overlooked issues in the growing literature of EEG-to-fMRI synthesis.

\end{abstract}

\section{Introduction}

Simultaneous electroencephalogram and functional magnetic resonance imaging (EEG-fMRI) is a valuable multimodal tool in human neuroscience, due to the complementary information about brain function provided by the two imaging modalities \citep{Jorge2014,Warbrick2022}. While EEG provides a direct measure of neuronal activity with high temporal resolution, its spatial specificity and coverage are limited due to volume conduction of electric currents through the brain, skull, and scalp. In contrast, fMRI provides exquisite spatial resolution with full brain coverage, despite being an indirect measure of neuronal activity based on blood-oxygenation-level-dependent (BOLD) changes reflecting neurovascular coupling mechanisms. The integration of EEG and fMRI from simultaneous recordings was initially motivated by the need to improve the localization of epileptic brain networks generating the abnormal electrical discharges observed in EEG. It was then extended to study healthy brain function, with the first studies aiming to map BOLD-fMRI activity associated with well-known EEG phenomena such as the alpha rhythm. Although such EEG-informed fMRI approaches have been the most common, a variety of asymmetric and symmetric integration techniques have been employed to analyze simultaneous EEG-fMRI data \citep{Abreu2018}. 

Recently, a growing interest has developed in the problem of learning models of EEG features able to predict brain activity patterns measured using BOLD-fMRI, the so-called problem of EEG-to-fMRI synthesis. While such predictive power would strengthen the evidence of the electrophysiology correlates of various phenomena observed using BOLD-fMRI \citep{Murta2015}, the main motivation for the growing literature addressing this problem of cross-modal reconstruction has been the design of EEG-only neurofeedback (NF) protocols targeting brain regions or networks that are more clearly captured by fMRI \citep{Fleury2023}. In particular, NF during the execution of motor imagery (MI) tasks has been explored for neurorehabilitation of stroke patients, by promoting brain plasticity and recovery of lost or impaired motor function \citep{Vourvopoulos2019}. Typically, NF signals are based on the EEG sensorimotor rhythm (SMR), which is a marker of activity originating in the sensorimotor brain areas, characterized by a reduction in alpha ($\sim$8-12 Hz) and beta ($\sim$13-30 Hz) power, known as event-related desynchronization (ERD) \citep{Pfurtscheller1999,Pfurtscheller2006}. Unfortunately, it is estimated that the SMR, which is also used to control brain computer interfaces (BCIs), cannot be measured in approximately 15-30\% of the general population, a phenomenon known as BCI illiteracy. In other NF applications, the goal is to modulate the activity of deep brain regions such as the amygdala or the insula, for which the EEG has minimal if any sensitivity. In contrast, the activity of these brain regions, as well as the somatomotor network, can be robustly measured using fMRI. However, it carries considerably higher costs, as well as safety, portability, and accessibility limitations. Hence, it would be of great interest to develop EEG-only NF protocols that could target brain activity patterns measured using fMRI. 

With this motivation, in a seminal study \cite{MeirHasson2014} used penalized linear regression based on EEG spectral power from a single electrode to obtain an EEG Finger-Print (EFP) of the BOLD-fMRI activity of the amygdala during a neurofeedback task. The authors further demonstrated that the EFP was significantly more predictive than conventional EEG theta/alpha activity. In a series of follow-up works, the authors expanded their approach from a subject-specific to a group model \citep{MeirHasson2016}, and further showed that it was effective in helping subjects downregulate their amygdala activity compared to a sham group, ultimately improving implicit emotion regulation \citep{Keynan2016}. Inspired by the EFP approach, others developed similar predictive models for different brain regions in a diverse set of neuromodulation applications \citep{Singer2023,Orborichev2023,Rudnev2021}. In \cite{Cury2020}, the authors moved beyond the single electrode EFP approach and developed a sparse regression model utilizing multi-channel EEG time-frequency features to predict EEG-fMRI NF scores from EEG signals during motor imagery. \cite{Simoes2020} employed random forests instead of linear regression, as well as a wide selection of EEG-derived features beyond spectral power, to predict the BOLD-fMRI activity of the facial expression processing network. The few works involving resting-state activity addressed different questions, demonstrating the ability to relate whole-brain EEG and fMRI connectomes, or the ability of EEG microstates to predict fMRI functional connectivity states \citep{Deligianni2014,Abreu2021,AlZoubi2021}. More recently, the problem has attracted interest from a broader audience and deep learning architectures such as convolutional neural networks (CNNs), generative adversarial networks (GANs), and transformers, have been proposed for synthesizing fMRI from concurrent EEG \citep{Calhas2020,lanzino2024,dagaev2024,Liu2019,Liu2023,Afrasiyabi2025,li2024_neurips, li2024, Yao2025}, while several others await peer review \citep{Calhas2022, Calhas2023,semenkov2024,Kovalev2022,roos2025,Liu2022,Donoso2025}. 

While many of these approaches show promise, the ability of EEG models to predict concurrent BOLD activity remains to be fully demonstrated. It is not clear, for example, to what extent spontaneous BOLD fluctuations can be predicted in comparison to task-evoked activity, or how models would perform for the same subject on a different day, which is of particular importance when considering NF interventions. In the present work, we acquired simultaneous EEG-fMRI data from a group of healthy volunteers in two separate sessions roughly two weeks apart, while they performed two MI tasks and during resting state. We analyzed the fMRI data to identify the activated SMN in each subject and condition, and we then trained models to predict its BOLD-fMRI activity from concurrent EEG signals. We chose interpretable distributed-lag linear models of multi-channel time-frequency EEG power, and we assessed their statistical significance and generalizability.

\section{Methods}
In this section, we first describe the dataset, including the participants, experimental protocol, and data acquisition details, as well as the EEG and fMRI data analysis pipelines used to extract the signal time series subsequently used for EEG-fMRI modeling. The final subsection describes the modeling approach, including model definition, estimation, selection, and validation.

\subsection{Participants and Experimental Protocol}

\begin{figure*}[t]
  \centering

  \begin{overpic}[width=\textwidth]{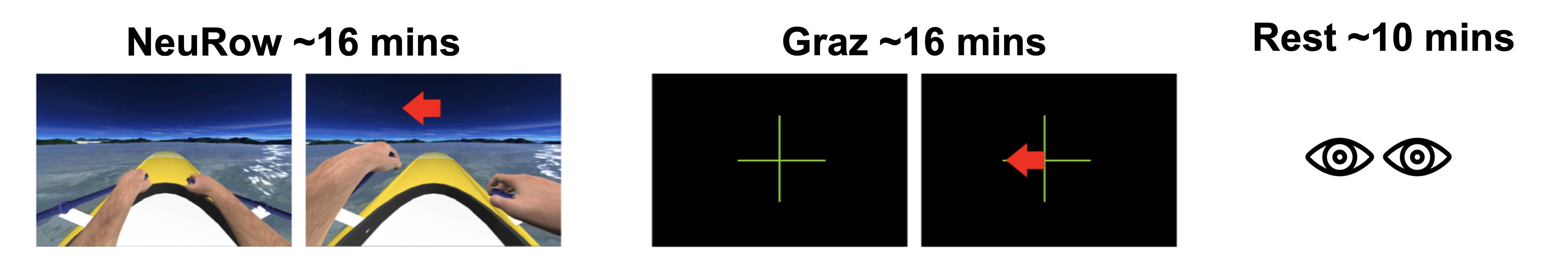}
        \put(1,15){\textbf{A}} 
    \end{overpic}\vspace{0.6em}

  \begin{overpic}[width=\textwidth]{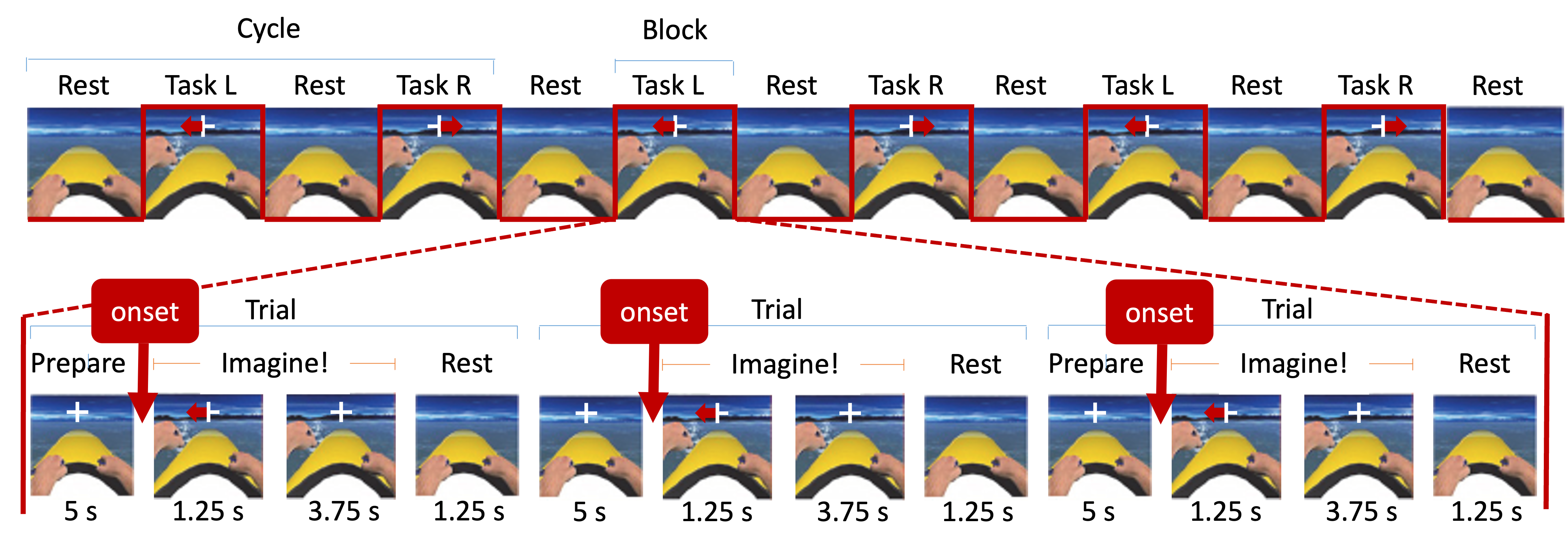}
        \put(1,33){\textbf{B}}
    \end{overpic}\vspace{0.6em}

  \caption{\textit{Experimental protocol.} (A) Illustration of the NeuRow, Graz and Rest scenarios, as presented to the participant in the screen. (B) Structure of one run (top) and one task block (bottom), for the NeuRow scenario. Blocks of left and right arm MI are separated by blocks of rest. Each trial is initiated with a preparation period and ends with a brief rest before the next trial. For the Graz scenario, the protocol timings are identical but only the arrow is presented to the subject as visual stimulus.}
  \label{fig:protocol}
\end{figure*}

Fifteen healthy volunteers (7 females, 8 males; mean age: 24.4 ± 2.7 years) participated in this study. The study was approved by Hospital da Luz Ethics Committee, and all participants provided written informed consent. For each subject, simultaneous EEG-fMRI data were acquired in two separate sessions approximately 2 weeks apart.

In each session, subjects were studied during the performance of two motor imagery tasks (Graz and NeuRow) and during a resting-state period (Rest). One task involved motor imagery only based on the Graz paradigm (Graz) \citep{Pfurtscheller2003}, while the other task also included observation of a first-person perspective of two virtual avatar arms on a boat with two paddles (NeuRow) \citep{Vourvopoulos2016,Nunes2023}. In both tasks, participants were instructed to imagine the kinesthetic experience of rowing using either their left or right arm, over blocks of three consecutive trials each, interleaved with rest. In Graz, a simple directional arrow served as the cue (left or right) against a black background with a white fixation cross. In NeuRow, the same arrow and fixation cross were presented, but this time superimposed on the virtual avatar arm displaying the instructed movement. Stimulus presentation was implemented using Unity, a game engine software, and was presented in synchronization with the EEG and fMRI acquisitions using NeuXus \citep{Legeay2022,Caetano2023}. The tasks were performed over 3 runs. The time structure of a run is illustrated in Fig.~\ref{fig:protocol}. Each run comprised 3 cycles, alternating 1 block of left arm imagery and 1 block of right arm imagery with rest blocks. Each block included 3 trials alternated with rest. Each trial was indicated by a fixation cross, and consisted of 5 s preparation followed by 1.25 s of arrow cue and 3.75 s of imagery. During rest periods, a simple black screen (Graz) or a static avatar (NeuRow) were presented. In each session, subjects performed a total of 54 trials (27 for each arm) over $\sim$17 mins per task. During Rest acquisitions, participants were instructed to lie still for 10 min while keeping their eyes open in dim light.

\subsection{EEG-fMRI Data Acquisition}
MRI scans were acquired using a 3T Siemens Vida system with a 64-channel-receive head radio-frequency coil. Functional images were acquired using a 2D Echo-Planar Imaging (EPI) sequence (TR = 1260 ms; TE
= 30 ms; flip angle = $70^{\circ}$; voxel size = 2.2 mm isotropic; in-plane acceleration with GRAPPA factor
2; simultaneous multi-slice with SMS factor 3; 60 axial slices). Structural images were acquired using a T1-weighted magnetization-prepared rapid gradient echo (MPRAGE) sequence (TR = 2300 ms; TE
= 2.98 ms; voxel size = 1.0 mm isotropic). 
EEG data were collected using a 32-channel MR-compatible EEG system (Brain Products GmbH, Germany), including 31 EEG channels and 1 electrocardiogram (ECG) channel. The EEG electrodes were placed on the head according to the 10-20 system, and the ECG electrode was placed on the back. The data were acquired using BrainVision Recorder (Brain Products, Germany) at a sampling rate of 5 kHz.

\begin{figure}[htbp]
    \centering
  \begin{overpic}[scale=0.9,trim={0 0 0 1cm}]{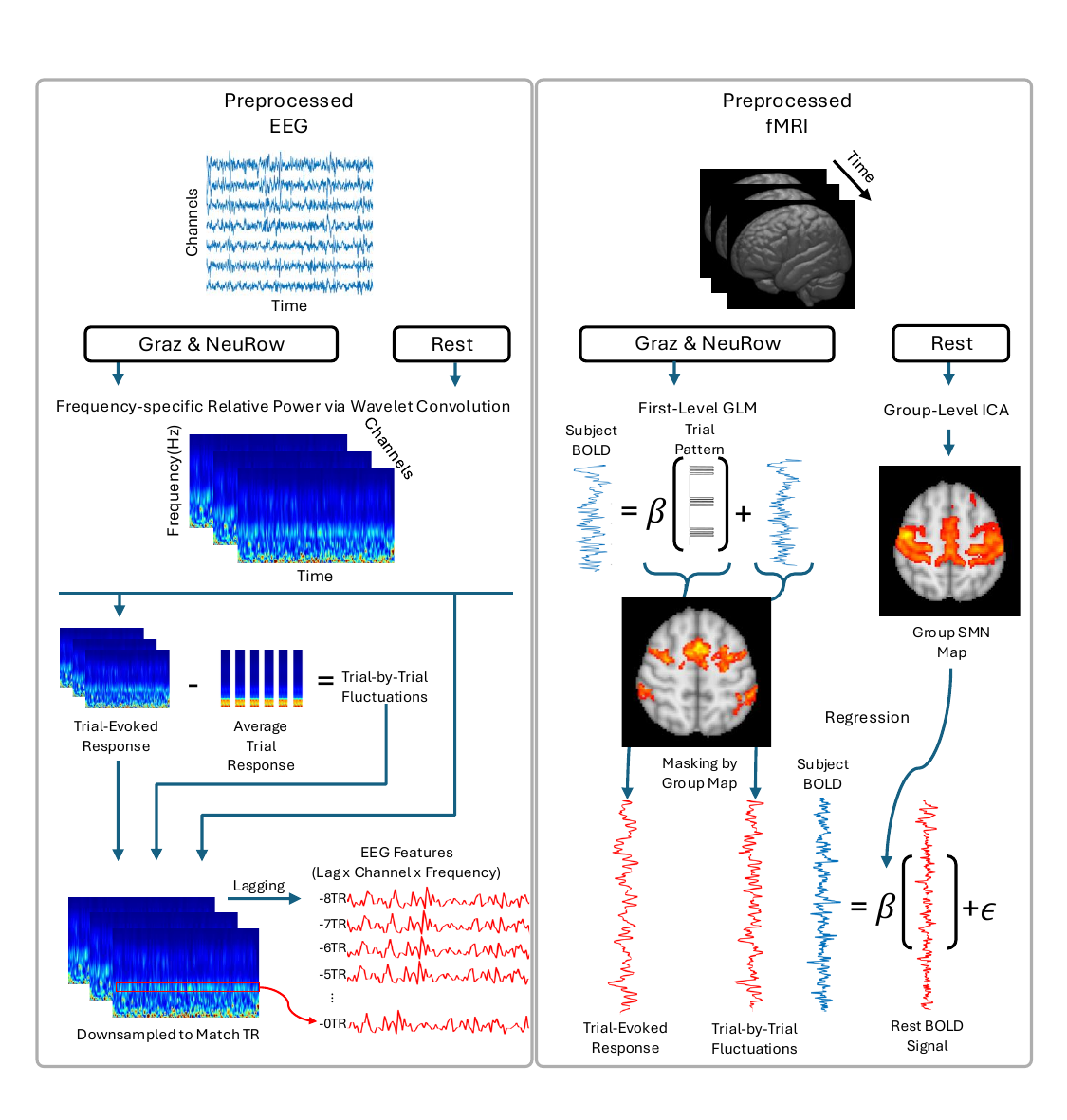}
        \put(5,95){\textbf{A}} 
        \put(52,95){\textbf{B}} 
    \end{overpic}\vspace{0.6em}
    \caption{\textit{Feature and Signal Extraction Pipelines.} (A) Relative power timeseries are extracted from EEG for each channel and frequency bin. From here task data is treated as the TE response, from which we subtract the average trial response centered on the trial stimuli times to produce TBT fluctuations. Finally, TE, TBT, and Rest timeseries are resampled to match the fMRI TR, before being shifted and stored in a matrix to serve as lagged regressors in our downstream models. (B) Task fMRI data is decomposed into TE responses and TBT fluctuations by first-level GLM analysis, and then masked by a group task-activation map. The rest data is submitted to a group-level ICA decomposition, and SMN IC selection. Finally, the first level of dual regression extracts a subject-specific SMN BOLD timecourse which is analogous to those found in the case of task data.}  
    \label{fig:pipeline}
\end{figure}

\subsection{EEG-fMRI Data Analysis} \label{fmri_preproc}

The data analysis pipeline is illustrated in Fig.~\ref{fig:pipeline}. Each step is described in detail in the following sections.

\subsubsection{fMRI Data Preprocessing}
Functional images were preprocessed using FMRIB Software Library (FSL) \citep{Smith2004}. A standard pipeline was followed, as described in the literature \citep{Jenkinson2002}, including: distortion correction using a fieldmap image, motion correction and realignment to the middle volume using rigid body transformation, removal of non-brain structures, high-pass temporal filtering (0.01 Hz cutoff frequency), spatial smoothing (3.3 mm full width at half maximum (FWHM) Gaussian kernel), co-registration to the structural image, and normalization to the Montreal Neurological Institute 152 (MNI152) standard space. 

For resting-state data, an additional denoising step was performed, in which the following nuisance time series were regressed out of the data: the six estimated motion parameters (MPs), motion outliers (MOs) identified as volumes for which the RMS intensity difference relative to adjacent volumes was above the 75th percentile  plus 1.5 times the interquartile range, and the average white matter (WM) and cerebrospinal fluid (CSF) signals obtained from masks defined by segmentation of the structural image.

\subsubsection{Extraction of fMRI Signals}

As illustrated in Fig.~\ref{fig:pipeline} - B, for each motor imagery task (Graz and NeuRow), subject, session and run, the preprocessed fMRI data were submitted to a first-level voxelwise general linear model (GLM) analysis using FSL's tool FEAT. The GLM was defined by convolving the square waveform of the task paradigm, for left and right-arm trials, with the canonical haemodynamic response function (HRF), as given by a double-gamma function with an overshoot at 6 seconds relative to onset. Additionally, its temporal derivative, the six MPs and the MOs were included as confounding regressors. Whole-brain images representing task-evoked activations were derived by fitting the GLM to each voxel's BOLD time series. To identify the group brain activation patterns associated with each task, a group-level mixed-effects GLM analysis was performed on the parameter estimates of the regressor of interest, and one-sample t-tests were converted to z-statistics and thresholded with a voxelwise threshold of $z > 3.1$, and an FWE-corrected cluster-level threshold of $p < 0.05$. 

The thresholded z-statistic images obtained for each subject, session, and run were binarized and then multiplied by a binary mask of the group-level activation map to obtain subject-specific task-related SMN maps. This was performed to ensure that each map was task-relevant, by removing voxels that showed no significant activation on average. A BOLD time series representative of the trial-evoked activity (TE) was obtained by averaging the BOLD signals across voxels within each map for each task, subject, session, and run. A BOLD time series representative of the trial-by-trial fluctuations (TBT) was obtained by averaging the residuals of the voxelwise GLM across voxels within each map for each task, subject, session, and run. Both the TE and TBT time series were converted to percent signal change with respect to the baseline, which was defined as the temporal mean of the BOLD signal in the period before the first trial in each run. After standardization of each run, the 3 runs were concatenated for each session.

The preprocessed resting-state fMRI data were subjected to group Independent Component Analysis (ICA) using FSL's tool MELODIC. The number of independent components (ICs) was set to 10, as a compromise between under-fitting and over-fitting \citep{Wang2015}. The z-statistic maps of the resulting ICs were thresholded at $z > 3.1$ and the SMN was identified as the IC yielding the maximum Dice coefficient between its map and Yeo’s template of the corresponding canonical RSN \citep{Yeo2011}. The first linear regression in dual regression was then performed to obtain the corresponding BOLD time series in each subject and session. This was converted to percent signal change with respect to the baseline, which was defined as the temporal mean of the BOLD signal over the whole run.

\subsubsection{EEG Data Preprocessing}
EEG data were preprocessed in MATLAB \citep{MATLAB} using code developed in-house based on tools from the open-source toolbox EEGLAB \citep{Delorme2004}. First, the gradient artifact (GA) was removed using moving average template subtraction. Additionally, an optimal basis set (OBS) extracted with principal component analysis (PCA) was used to fit and remove the residuals, implemented in the EEGLAB-plugin FMRIB \citep{niazy2005}. The ECG R-peaks were detected using a Long Short-Term Memory (LSTM) network, and manually corrected when needed, and the EEG data were then corrected for the pulse artifact (PA) using a combination of ICA and the optimal basis set (OBS) technique \citep{Caetano2023,Abreu2016}. Following MR artifact correction, EEG data were downsampled to 250 Hz and bandpass filtered from 1 to 40 Hz. Bad EEG channels were removed, all EEG channels were re-referenced to their average, and subjected to ICA to identify and remove physiological artifacts such as muscle activity and eye movements. Finally, interpolation of burst activity was performed using Artifact Subspace Reconstruction (ASR) \citep{Chang2020}.

\subsubsection{Extraction of EEG Features}
EEG feature extraction, illustrated in Fig.~\ref{fig:pipeline} - A, was performed using Brainstorm \citep{Tadel2011}. The time-varying power spectra were extracted from each of the 31 EEG channels (excluding ECG) by application of the continuous wavelet transform using a complex Morlet mother wavelet with a central frequency of 1 Hz and a time resolution of 3 s \citep{Bertrand2000,Pantazis2005}. We extracted wavelet coefficients for each whole number frequency in the range 1-40 Hz and approximated the time-varying relative power by squaring the coefficients and dividing by the total power across the spectrum at each sample in time.

For the TBT conditions, an additional step was performed of removing the average trial-evoked response for both left and right trials. Practically, this process involved epoching the EEG time-frequency data in each channel over 10 seconds windows centered on each trial onset and averaging across trials in each run to produce a stereotype of the subject's response to the task. This stereotype was then subtracted from the relative power time series at each trial to leave only trial-by-trial fluctuations.

We temporally downsampled these time series using an anti-aliasing low-pass filter to match the sampling rate of the BOLD signal. For the tasks Graz and NeuRow, we standardized EEG relative power regressors for each run before concatenating them across the 3 runs of each session. Finally, to account for variable haemodynamic delays, we shifted the resulting EEG signals forward relative to the BOLD signal by 8 time lags, covering a physiological range of approximately 0-10 s, $\{0,1,2,\ldots,8 \}$TRs.

\subsection{EEG-fMRI Modeling}
To summarize, for each subject and session, BOLD-fMRI signals and corresponding EEG features were extracted in the five following conditions: 1) Graz - TE; 2) NeuRow - TE; 3) Graz - TBT; 4) NeuRow - TBT; and 5) Rest. While conditions 1) and 2) correspond to brain activity evoked by task trials, conditions 3) to 5) correspond to spontaneous fluctuations in brain activity, either from trial to trial during the execution of a task (3 and 4) or during rest (5). In the following sections, we describe the modeling decisions and estimation techniques employed to generate informed time series predictions of the BOLD signal for each subject, session and condition.

\subsubsection{Model Definition} \label{eegtofmri}
To accommodate the temporal lag of the BOLD signal relative to the EEG signal, we constructed a \textit{distributed lag model} (DLM) i.e. a linear model that includes individual regressors for each EEG channel and frequency, as well as lagged copies of each of these time series. The input data is a tensor, $X \in \mathbb{R}^{T \times C \times F}$, where $T, C$, and $F$ represent the duration of the EEG signal ($T=804$), the number of channels ($C=31$), and the number of frequencies ($F=40$), respectively. Let $y \in \mathbb{R}^T$ denote the true BOLD signal derived as in ~\ref{fmri_preproc}. Flattening our input tensor, $X$, to form a design matrix, $\mathcal{X} \in \mathbb{R}^{T \times R}$ where $R = C \times F$, allows us to express this model as follows,

\begin{equation} \label{eq1}
y =  \beta \mathcal{X} + \epsilon = \beta_0 + \sum_{r \in [R]} (\beta_{r}^{(0)} L^0 \mathcal{X}_{r} + \beta_{r}^{(1)} L^1\mathcal{X}_{r} + \ldots + \beta_{r}^{(M-1)} L^{M-1}\mathcal{X}_{r}) + \epsilon
\end{equation}

where the sum is taken over the total number of regressors across all channels and frequencies, $R$, and $L^n x_t = x_{t-n}$ is the lag operator of order $n$. Individual coefficients $\beta_{r}^{(n)}$ account for the effects of each of these regressors, at the specific lag $n$ up to a maximum specified lag order, $M-1$. The errors, $\epsilon$, are assumed to be normally distributed and uncorrelated in time. Based on the range of previously reported haemodynamic lags, as well as our preliminary massive univariate correlation analysis presented in Fig.~\ref{fig:massive_univariate}, we include the EEG time sample aligned with the current fMRI sample we are attempting to predict as well as the previous 8 time points (fMRI acquisitions), corresponding to approximately 10 s ($M=9$).

\subsubsection{Model Estimation}

We opted for the Sparse Group Lasso (SGL) \citep{Simon2013} formulation due to the natural hierarchical structure of our features i.e. several frequency or lag regressors were derived from the signal on the same EEG channel. The Sparse Group Lasso problem is formulated as follows,

\begin{equation} \label{eq2}
\beta^{*} = \argmin_{\beta} \frac{1}{2n}||y - \sum_{c \in C} \beta_{(c)} \mathcal{X}_{(c)}||_2^2 +\lambda (1-\alpha) \sum_{c \in C} \sqrt{p_c} ||\beta_{(c)}||_2 + \lambda \alpha ||\beta||_1
\end{equation}

The tradeoff between $\ell_1$ and $\ell_2$ regularization penalties are controlled by $\alpha$, while $\lambda$ controls the overall strength of regularization. Scaling the group penalty by the length of $\beta_{(c)}$, $\sqrt{p_c}$, ensures that distinct group sizes do not influence the optimization. Model parameters are estimated using a variant of the Fast Iterative-Shrinking Threshold Algorithm (FISTA) for its efficiency \citep{Moe2019}.

\subsubsection{Cross-validation}

\begin{figure}[htbp]
    \centering
  \begin{overpic}[scale=0.9,trim={0 0 0 1cm}]{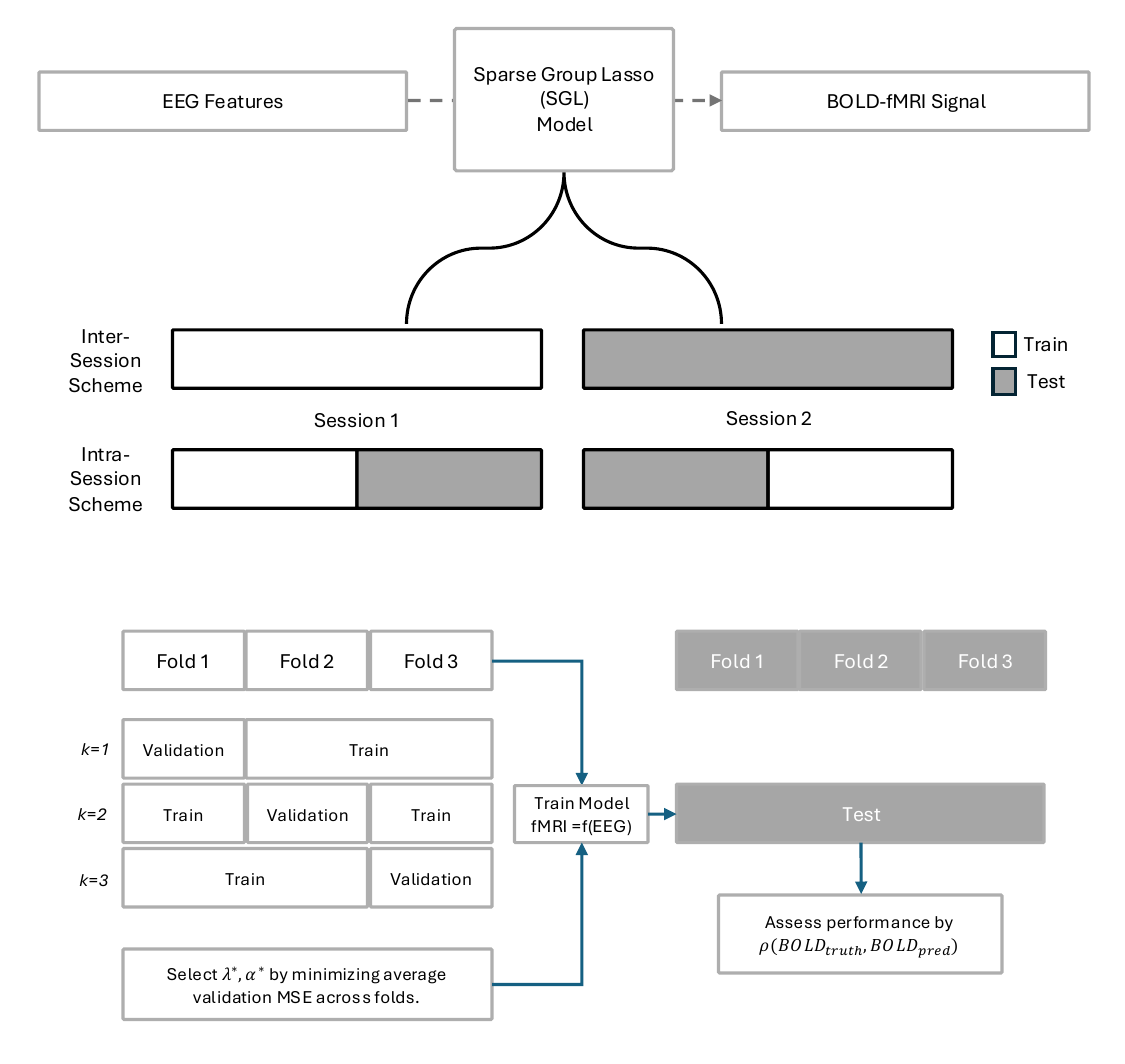}
        \put(5,75){\textbf{A}} 
        \put(5,40){\textbf{B}} 
    \end{overpic}\vspace{0.6em}
    \caption{\textit{Nested Cross-Validation Schemes.} (A) Two train-test splitting schemes were used: inter-session (keeping the two sessions separate) and intra-session (mixing the two sessions). (B) For hyperparameter optimization, a block cross-validation scheme was used, by splitting the training dataset into 3 folds. A model is then trained using the entire training dataset and the selected hyperparameters and then applied to the held out test dataset and the Pearsomn correlation between the model prediction and the true BOLD signal is computed to assess model performance.}  
    \label{fig:crossval}
\end{figure}

The cross-validation scheme is illustrated in Fig.\ref{fig:crossval}. 

We considered two schemes for splitting the data for training and testing: inter-session scheme, whereby one of the sessions is used for training and the other one for testing; and intra-session scheme, whereby the data from the two sessions are merged by concatenating the first half of session 1 with the second half of session 2, and the first half of session 2 with the second half of session 1, such that the training and testing datasets contain half the samples from both sessions. In both schemes, the data split is 50\% train / 50\% test, yielding two parcellations in each case. The performance of the models is computed as the averaged over both parcellations (unless otherwise stated). We use the intra-session scheme to investigate potential leakage effects when using test and training data acquired in the same session. 

We performe block \textit{k}-fold cross validation ($k=3$) to optimize the regularization hyperparameters, $\alpha$ and $\lambda$. The choice of a block cross-validation scheme preserves the ordering of samples in time, which is important not to contaminate the training and validation partitions by allowing serially correlated samples to appear in training and validation partitions simultaneously. We utilize Bayesian optimization \citep{bayesianoptimization} to preferentially sample the validation mean squared error (MSE) between our true and predicted fMRI time series, using the Upper Confidence Bound to acquire new samples with $\kappa = 0.1$ i.e. preferring exploitation, not exploration. Optimal hyperparameters are chosen over a gaussian process smoothed estimate of the validation loss surface. A model is then trained using the entire training dataset and the selected hyperparameters and then applied to the held out test dataset. The Pearson correlation between the predicted and true BOLD time series of the held out dataset is computed and used as the discriminating statistic to assess the generalization performance of our learned model.

\subsubsection{Statistical Assessment of Model Prediction}
To ascertain the statistical significance of the model's predictions, we utilize the \textit{method of surrogates} \citep{Lancaster2018}. This method involves generating a large number of surrogate samples of BOLD time series consistent with an appropriate null hypothesis and applying the model estimation procedure to these surrogate time series to obtain a null distribution associated with our null hypothesis. The p-value of the chosen discriminating statistic is then computed under the null distribution.

Our null hypothesis is that the BOLD time series cannot be explained by any linear combination of EEG features at any lag. \textit{Fourier Transform surrogates (FT)} are commonly used to generate surrogate sample time series that preserve the autocorrelation function and power spectral density of the original time series. A model trained with unperturbed EEG time series and FT surrogates of the BOLD time series is consistent with the null hypothesis. However, this surrogate generation procedure may admit false positives if our initial BOLD time series is produced via a non-Gaussian stochastic process, or if it is non-stationary. To mitigate non-Gaussianity, we utilize the variant \textit{Iterative Amplitude-Adjusted Fourier Transform (IAAFT)} surrogates \citep{Lancaster2018}, which avoids false positives due to non-Gaussianity by adjusting the amplitude distribution of the final surrogate to better match the initial starting time series. To mitigate non-stationarity, we preemptively test each BOLD time series using the Augmented Dicky-Fuller test (ADF) to ensure stationarity ($p < 10^{-5}$ for all time series). 

For each BOLD surrogate time series, a model is estimated based on the unperturbed EEG time series, and its generalization performance is measured using Pearson correlation between predicted and true surrogate time series (as for the true BOLD time series). This correlation represents a realization of the distribution of the null hypothesis.To assess the statistical significance of the model's prediction for each subject, session, and condition, a null distribution is obtained by generating 100 surrogates. P-values are computed as the number of surrogate models for which the Pearson correlation is greater than or equal to the Pearson correlation between predicted and true BOLD signals, normalized by the number of surrogates. 

\subsubsection{Comparison with Sensorimotor Rhythm and Massive Univariate Correlations}

To further assess the value of the predictions obtained with the learned models, we compared them with two reference models: a conventional approach using the somatomor rhythm (SMR), which was not learned from the data but rather informed by modern BCI literature \citep{yuan2014}; and Massive Univariate (MUC) models. 

We defined the SMR model by averaging activities from C3 and C4 channels in the alpha and beta bands (8-30 Hz), and shifting it relative to the BOLD signal by 6.3 s (corresponding to a canonical hemodynamic delay). 

We defined the MUC model by computing the Pearson correlation between each EEG regressor (frequency-specific power time series from each EEG channel and for each haemodynamic lag) and the concurrent BOLD time series, for each condition, subject and session.

\section{Results}

In this section, we first present the results of the fMRI analysis leading to the mapping of the SMN and extraction of the respective BOLD activity time courses. We then present the results of the model prediction, including the subject-level statistical evaluation against null models as well as the group-level comparisonwith the two control models. In all cases, the results are presented for both the inter-session and intra-session train-test schemes.

\subsection{fMRI SMN mapping and BOLD time series extraction}

\begin{figure}[htbp]
    \centering

    \includegraphics[scale=0.5]{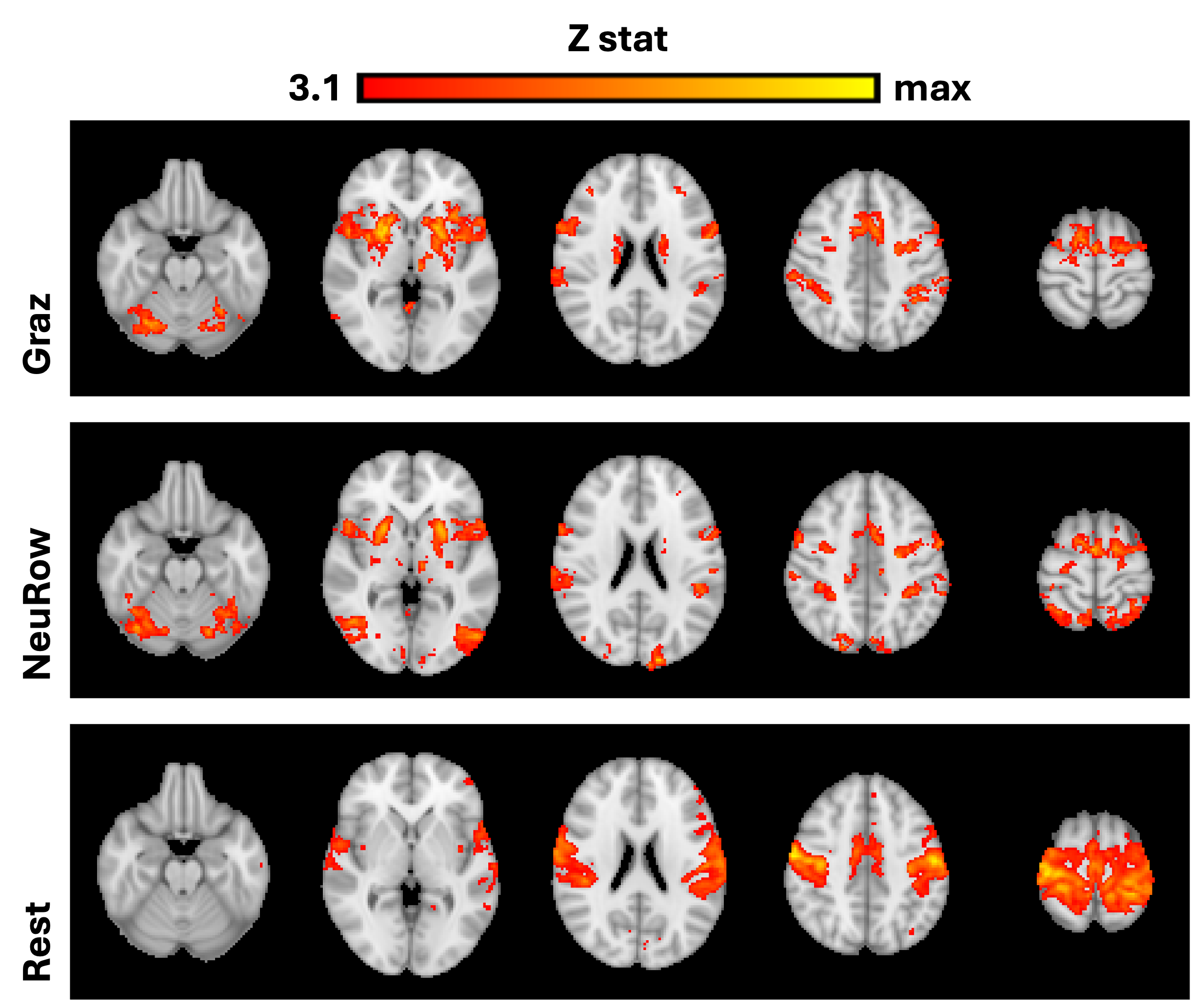}

    \caption{\textit{Group fMRI SMN maps.} SMN maps identified for both tasks as the average motor imagery activation across subjects and sessions (Graz, Neurow), and for rest by group ICA and subsequent identification of the SMN network by template matching (Rest). The Z statistic maps obtained after thresholding for statistical significance (color) are overlaid on the MNI anatomical brain template (gray).}  
    \label{fig:networks}
\end{figure}

\begin{figure}[htbp]
    \centering
    \begin{overpic}[scale=0.5,clip]{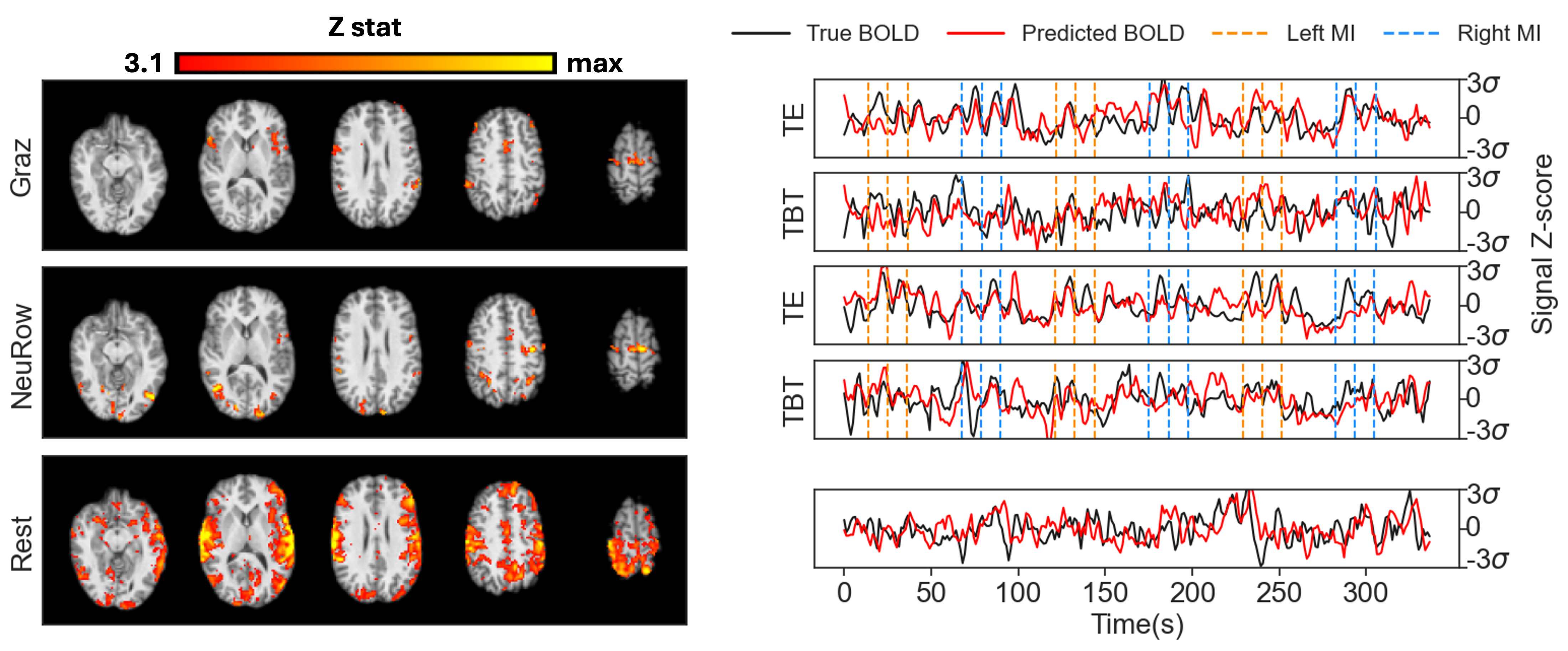}
        \put(1,40){\textbf{A}} 
        \put(45,40){\textbf{B}}
    \end{overpic}
    \caption{\textit{fMRI results for a representative subject} (A) fMRI SMN maps obtained for each task (Graz and NeuRow) and for resting state (Rest). (B) BOLD-fMRI time series (black) and respective model predictions (red), for both task-evoked (TE) and trial-by-trial (TBT) activities for the tasks (Graz - TE, Graz - TBT, Neurow - TE, NeuRow - TBT), with trial onset times indicated by the dashed lines (orange -- left, blue -- right).}  
    \label{fig:sub04}
\end{figure}

The group-level fMRI maps of the SMN obtained for the MI tasks (Graz and NeuRow) and the resting state (Rest) are shown in Fig.~\ref{fig:networks}. An illustrative example for a representative subject of the fMRI SMN maps for the three tasks, together with the respective BOLD time series obtained for the five conditions (Graz - TE, Graz - TBT, NeuRow - TE, NeuRow - TBT, and Rest), are shown in Fig.~\ref{fig:sub04}. To assess the conventional EEG SMR present in the dataset, the grand-average ERD obtained across all trials, subjects and sessions, for each MI task (Graz and NeuRow), as well as an illustrative example for a representative subject, are shown in Supplementary Material (Fig.~\ref{fig:erd}).

\subsection{EEG-to-fMRI Prediction: Comparison with SMR and MUC Models} \label{group_results}
\begin{figure}[htbp]
    \centering
    \includegraphics[width=0.5\textwidth]{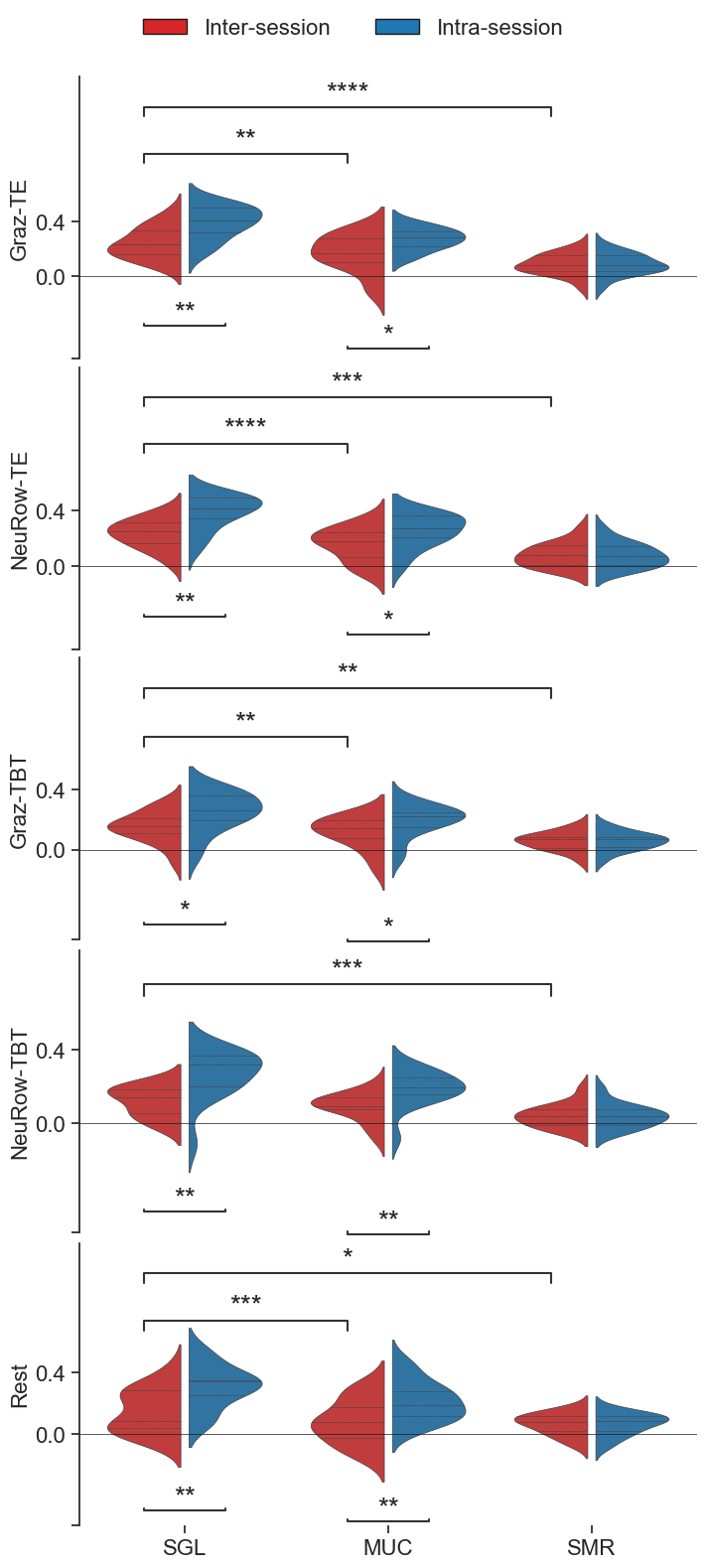}
    \caption{\textit{EEG-to-fMRI Prediction: Comparison of SGL to Reference Models.} (Left to Right) Test performance of Sparse Group Lasso (SGL), Massive Univariate Correlation (MUC), and Sensorimotor Rhythm (SMR) models in each condition, for both inter- and intra-session train-test paradigms. Each violin plot represents the distribution of test performance across subjects. Dashed and dotted lines indicate median and IQR limits, respectively. Wilcoxon Signed-Rank tests, with Benjamini-Hochberg FDR correction, were used to compare SGL to MUC and SMR as well as to compare each inter-session and the corresponding intra-session distributions. Significant results are indicated with * $\alpha \leq 0.05$, ** $\alpha \leq 0.01$, *** $\alpha \leq 0.001$, and **** $\alpha \leq 0.0001$.}
    \label{fig:task_significance}
\end{figure}

The group-level evaluation of the test performance of the learned models compared to the respective SMR and MU models is presented in Fig.~\ref{fig:task_significance}. 

Using Wilcoxon signed-rank tests, we found that the learned models achieved significantly better test correlations than the respective SMR and MUC models in all conditions. These results indicate that the learned models generally outperform conventional predictors, suggesting that we benefit from other sensor locations and frequency components beyond C3 and C4 alpha/beta activity predictors (SMR), and further that we benefit from learning a properly regularized model over all available EEG information (MUC). This observation is further evidenced by the frequency of the optimal hyperparameter values found for the learned models which is presented in Supplementary Material (Figure~\ref{fig:hyperparams}). 

We find similar hyperparameters for most subjects regardless of whether the models are trained under the \textit{inter}- or \textit{intra}-session paradigms, and the relatively low $\lambda$ values indicate that the optimal model fits rely on Ridge regression ($\ell_2$) as opposed to LASSO ($\ell_1$). In other words, models are not sparse but instead rely on all or most of the EEG regressors to produce a more accurate prediction. Further analysis of which channels and frequencies are the most informative predictors of simultaneous BOLD activity with respect to different conditions is presented in ~\ref{model_interpretation}. 

Critically, the intra-session train-test paradigm yielded significantly higher test correlations than the inter-session test paradigm for every condition using the SGL as well as the MU models, but not for the SMR. This finding indicates that leakage effects occur between train and test data when retrieved from the same session leading to an optimistic bias in model predictions, and that this effect is more pronounced when more diverse EEG information is integrated into the prediction.

\subsection{EEG-to-fMRI Prediction: Statistical assessment}
\begin{figure}[htbp]
    \centering
    \includegraphics[scale=0.35]{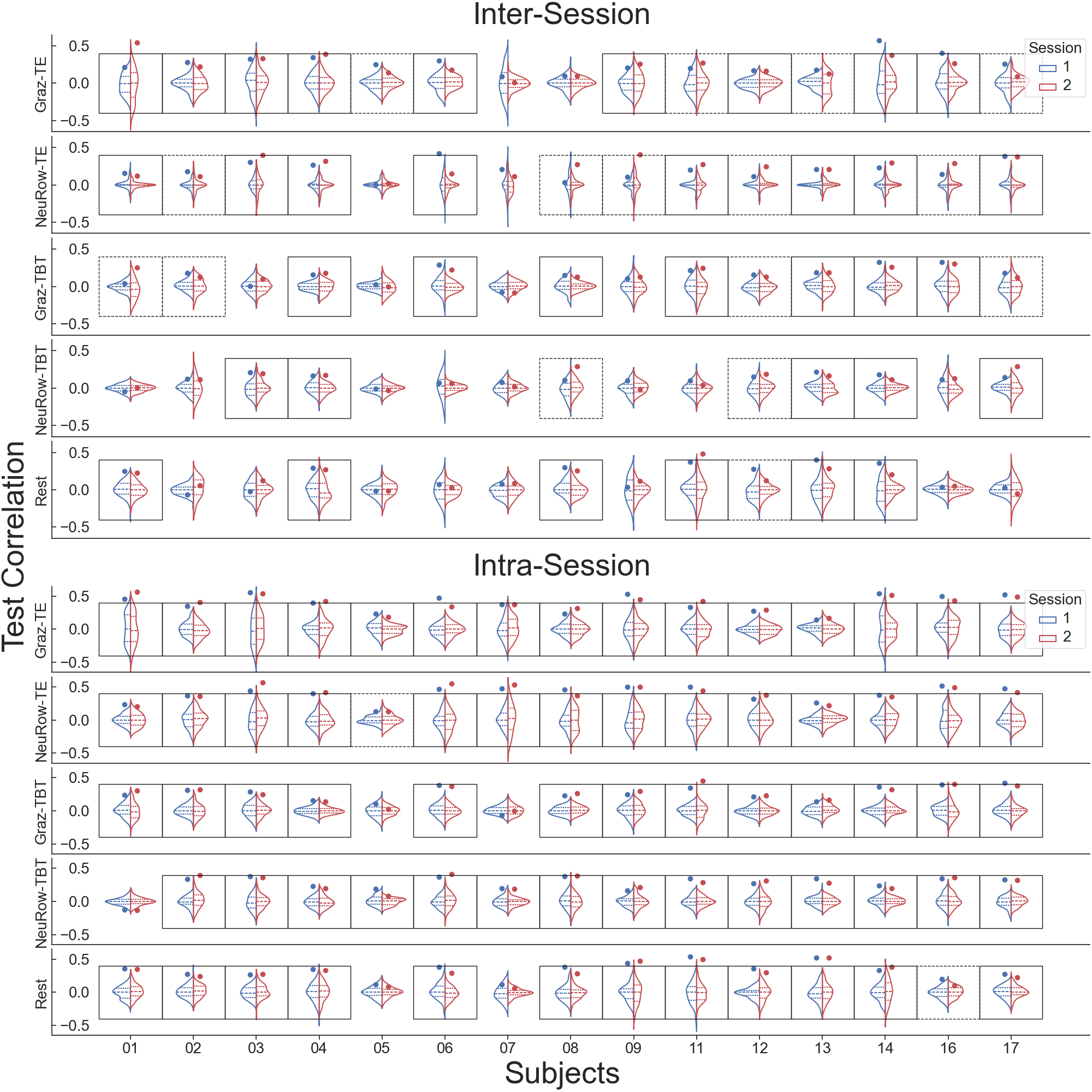}
    \caption{\textit{Test Performance of Sparse Group Lasso EEG-to-fMRI Models against Null Distributions.} Subject-specific model test performance (blue and red dots) compared to null distributions obtained by training using IAAFT surrogate fMRI timeseries (blue and red violin plots). Boxes with dotted or solid lines denote significance of one or both session-specific models compared to the null distribution ($\alpha = 0.05$). Results are shown for both intra- and inter-session train-test paradigms.}
    \label{fig:subject_significance}
\end{figure}

The test performance of the models learned with SGL to predict SMN BOLD-fMRI activity from concurrent EEG power features, in each subject and session, for each of the five conditions tested with both train-test paradigms, is presented in Fig.~\ref{fig:subject_significance}. Table~\ref{tab:summary} summarizes the number of subjects for which a significant prediction was achieved, for one or both sessions, within each condition and for both train-test paradigms. 

\begin{table}[h!]
  \caption{\textit{EEG-to-fMRI prediction statistical evaluation:} number of subjects for whom prediction was statistically significant for one session only (One), both sessions (Both), and at least one session (One or Both) (out of a total of 15 subjects), for each condition and for both inter-session and intra-session train-test paradigms.}
  \small
  \centering
  \begin{tabular}{l l c c c}
    \toprule\toprule
    \textbf{Train-Test Paradigm} & \textbf{Condition} & \textbf{One Session} & \textbf{Both Sessions} & \textbf{Total} \\
    \midrule

    \multirow{5}{*}{\textbf{Inter-Session}}
      & Graz--TE     & 4 / 15 & 9 / 15 & 13 / 15 \\
      & NeuRow--TE   & 5 / 15 & 8 / 15 & 13 / 15 \\
      & Graz--TBT    & 4 / 15 & 7 / 15 & 11 / 15 \\
      & NeuRow--TBT  & 2 / 15 & 5 / 15 & 7 / 15 \\
      & Rest         & 1 / 15 & 6 / 15 & 7 / 15 \\
    \midrule

    \multirow{5}{*}{\textbf{Intra-Session}}
      & Graz--TE     & 0 / 15 & 15 / 15 & 15 / 15 \\
      & NeuRow--TE   & 1 / 15 & 14 / 15 & 15 / 15 \\
      & Graz--TBT    & 0 / 15 & 13 / 15 & 13 / 15 \\
      & NeuRow--TBT  & 0 / 15 & 14 / 15 & 14 / 15 \\
      & Rest         & 1 / 15 & 12 / 15 & 13 / 15 \\
    \bottomrule
  \end{tabular}
  \label{tab:summary}
\end{table}

In each case, we compare the model discriminative statistic (Pearson correlation between predicted and true BOLD time series) to the respective null distribution, and indicate cases where a statistically significant difference was found. While a significant prediction was achieved for at least one session in most subjects in the TE conditions for both Graz and NeuRow tasks, this number was reduced in TBT of NeuRow and in Rest. Of particular note, subjects 5 and 7 produced almost no significant predictions in any condition, potentially due to BCI illiteracy. Also, consistently with the results in ~\ref{group_results}, the intra-session train-test paradigm yielded more optimistic results than the inter-session train-test paradigm, with significant predictions being achieved in every subject for at least one session in the TE conditions, again highlighting the importance of temporally separating collection of train and test data if one hopes to reuse the same mode for the same subject on a different day.

\subsection{EEG-to-fMRI Prediction: Model Interpretability} \label{model_interpretation}
To assess whether the learned models are physiologically meaningful, we exploited the interpretability of the linear models learned and analyzed the patterns of model coefficients, comparing these with the patterns of MUC coefficients.

\begin{figure}[htbp]
    \centering
    \begin{overpic}[scale=0.275]{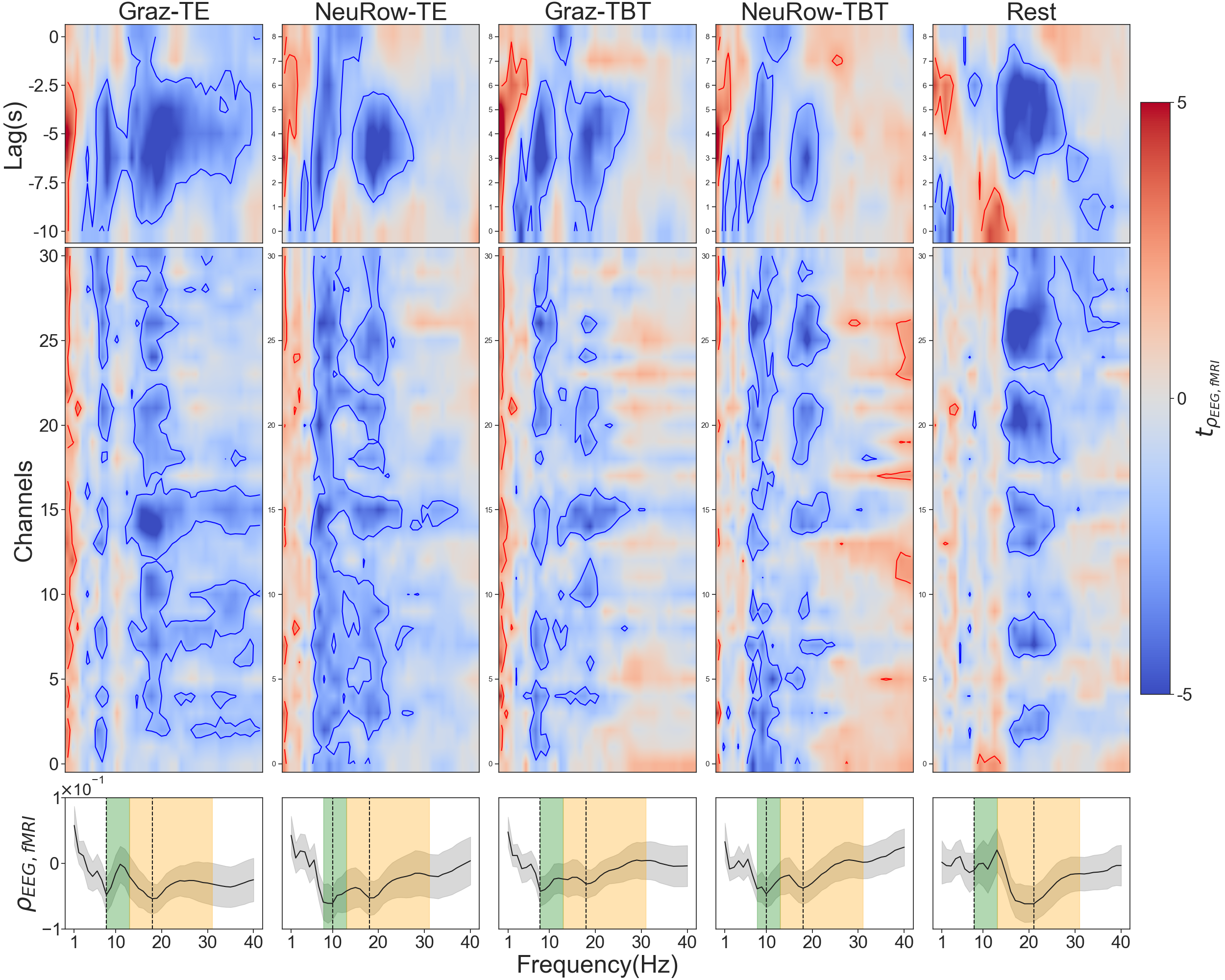}
        \put(1,78){\textbf{A}} 
        \put(1,58){\textbf{B}}
        \put(1,14){\textbf{C}}
    \end{overpic}
    \caption{\textit{EEG-fMRI Massive Univariate Correlation Coefficients.} 
    T-test values across subjects and sessions for each of the five conditions tested (Graz - TE, NeuRow - TE, Graz - TBT, NeuRow - TBT, Rest) with bilinear interpolation: (A) Heatmaps of t-statistics of maximum absolute correlation taken across channels, as a function of frequency and lag. (B) Heatmaps of t-statistics of maximum absolute correlation across lags, as a function of frequency and channel. Channels arranged from anterior to posterior brain regions from top to bottom. (C) Mean-across-channels curves of the maximum absolute correlation taken across lags, as shown in B. Green and orange distinguish frequencies in the alpha (7-12 Hz) and beta (12-30 Hz) bands. In (A) and (B), the red and blue contours indicate clusters of t-values with ($\alpha = 0.05$). In (C), the black lines indicate the negative peaks.}
    \label{fig:massive_univariate}
\end{figure}

The patterns of univariate temporal correlations between the SMN BOLD-fMRI activity and concurrent EEG power features, along frequencies, channels and lags, are presented in Fig.~\ref{fig:massive_univariate}, as t-test values across subjects and sessions. To simplify visualization, 2D heatmaps are generated to show patterns across frequencies x channels and frequencies x lags. Before aggregation, the maximum absolute correlation (preserving sign) is taken for each subject and session across lags (frequencies x channels) or across channels (frequencies x lags). The mean across channels of the maximum absolute correlations (frequencies x channels) are also analyzed as a function of frequency, showing two negative peaks, one in the alpha band and the other in the beta band. The identification of negative correlations in the alpha and beta bands, with a focus in central and parietal regions is partially consistent with the SMR. Additional analysis of the most predictive channels (for each lag and frequency) and lags (for each channel and frequency) are presented in Supplementary Material (Fig.~\ref{fig:massive_univariate_ext}). 

\begin{figure}[htbp]
    \centering
    \begin{overpic}[scale=0.275]{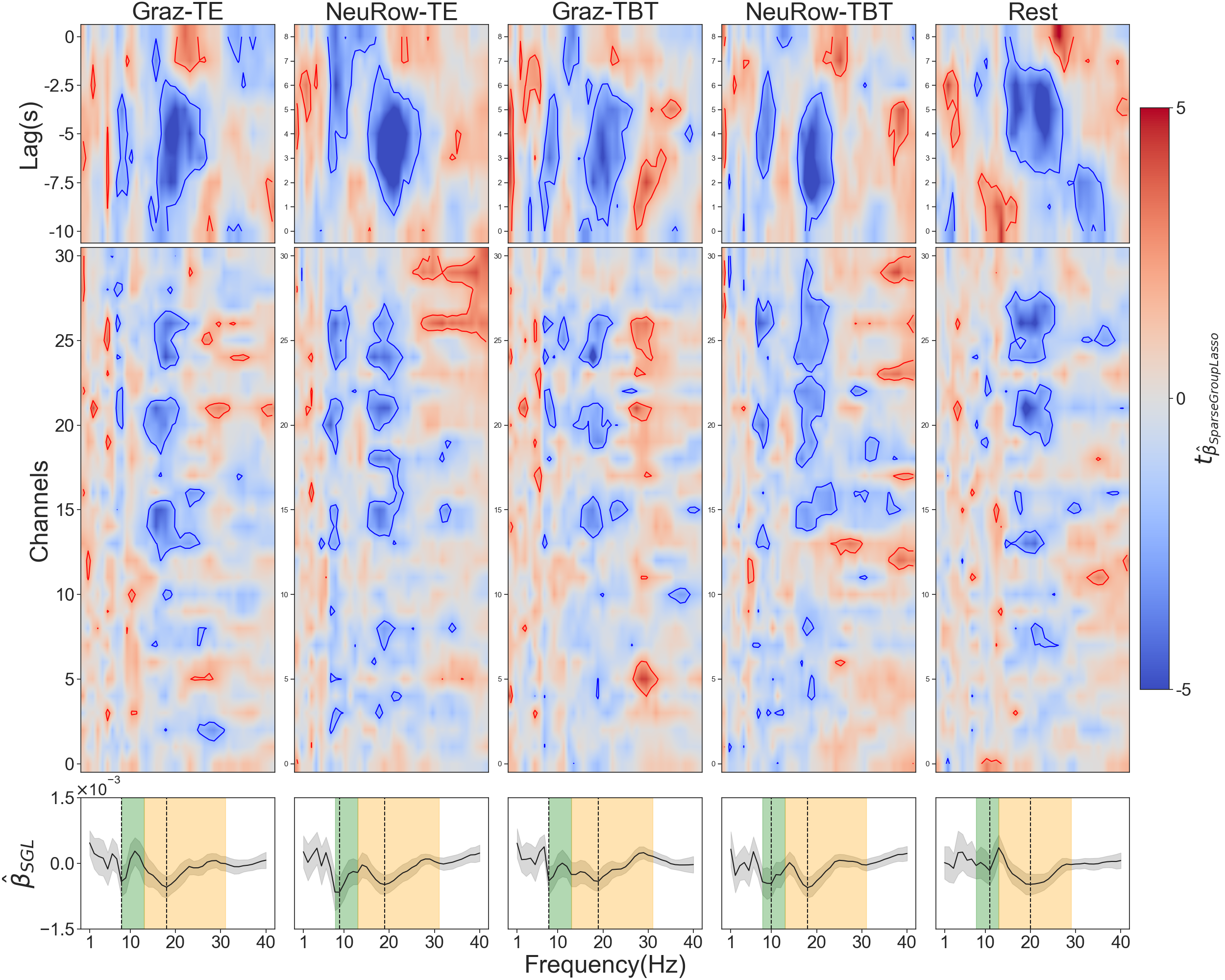}
        \put(1,78){\textbf{A}} 
        \put(1,58){\textbf{B}}
        \put(1,14){\textbf{C}}
    \end{overpic}
    \caption{\textit{EEG-to-fMRI Sparse Group Lasso Model Coefficients} 
    T-test values across subjects and sessions for each of the five conditions tested (Graz - TE, NeuRow - TE, Graz - TBT, NeuRow - TBT, Rest) with bilinear interpolation: (A) Heatmaps of t-statistics of maximum absolute coefficient taken across channels, as a function of frequency and lag. (B) Heatmaps of t-statistics of maximum absolute coefficient across lags, as a function of frequency and channel. Channels arranged from anterior to posterior brain regions from top to bottom. (C) Mean-across-channels curves of the maximum absolute coefficient taken across lags, as shown in B. Green and orange distinguish frequencies in the alpha (7-12 Hz) and beta (12-30 Hz) bands. In (A) and (B), the red and blue contours indicate clusters of t-values with ($\alpha = 0.05$). In (C), the black lines indicate the negative peaks. } 
    \label{fig:model_coefficients}
\end{figure}

Similar to the MUC results, the maximum absolute correlations (preserving sign) are t-tested across subjects and sessions and presented in Fig.~\ref{fig:model_coefficients}, arranged such that the EEG frequencies, channels, and hemodynamic lags that are important for predicting concurrent BOLD signals are clear. Recall from~\ref{eegtofmri} that a coefficient is estimated for each regressor in a $C \times F \times M$ tensor. To simplify the visualization of the results, we follow the same approach as for the MUC analysis in Fig.~\ref{fig:massive_univariate} and show 2D heatmaps, by averaging the coefficients across lags (frequencies x channels) or across channels (frequencies x lags). Again, an additional analysis was performed to identify the most predictive channels and lags. The results are presented in Fig.~\ref{fig:distributed_lag_ext}.

\section{Discussion}

We demonstrate that it is possible to predict both evoked and spontaneous BOLD activity of motor brain networks based on interpretable EEG models trained on EEG-fMRI data acquired from the same subject on a different day. The learned models are not only significantly better than conventional EEG SMN features in all conditions, but to varying degrees they exhibit a statistically significant predictive power in most subjects when compared with null models. Importantly, spontaneous trial-by-trial fluctuations during motor imagery tasks and resting-state activity can also be predicted even if with lower fidelity than task-evoked activity. Most critically, we provide evidence of statistical leakage effects between training and test data retrieved from the same acquisition session, leading to an optimistic bias in model predictions.

\subsection{Relation to literature}
Ours is the first study aiming to predict the BOLD activity of the motor network from concurrent EEG. Nevertheless, one other study addressed a very similar problem, that of predicting EEG-fMRI NF scores from EEG signals, employing a similar penalized linear regression approach \citep{Cury2020}. In contrast to this study, we employed a Sparse Group Lasso regularization, allowing us to achieve informative spatial sparsity while maintaining smoothness in the temporal and frequency domains. Moreover, we created appropriate null models to assess the statistical significance of the predictions in each subject. Most importantly, we went beyond task-evoked activity to investigate the ability of the models to predict even spontaneous fMRI signals, either across trials or during rest.

More generally, most previous works aiming to predict fMRI signals from EEG also leveraged the well-known relationship between EEG oscillatory activity and BOLD-fMRI activation \citep{Formaggio2010,Scheeringa2012,Murta2015}. However, most reduced either the spatial dimension (for example, the use of a single EEG electrode in EFP models as described in \cite{MeirHasson2014,MeirHasson2016}) and/or using only EEG SMR frequencies (alpha and beta) \citep{Cury2020}. Instead, we opted to preserve the full spatial and frequency resolution of EEG signals to allow our model to estimate relevant relationships in a data-driven fashion. Importantly, the model we employed strikes a balance between complexity and expressivity as compared to modern deep learning approaches to also allow for rigorous model validation and significance testing. In general, our work is the first on EEG to fMRI prediction that employs appropriate null models to assess the significance of model prediction. Moreover, ours is also the first work to directly assess potential leakage effects between training and test data when retrieved from the same session, a procedure that is adopted in the majority of previous works.

\subsection{Prediction significance}
The results of our surrogate data analysis demonstrate that the statistical significance of fMRI signal predictions from concurrent EEG signals depends strongly on whether we aim to predict task-evoked or spontaneous activities. In fact, it was possible to achieve significant predictions of task-evoked activity (Graz - TE and NeuRow - TE) for over 70\% of the subjects and sessions. However, this was reduced to about 50\% for spontaneous activities, either trial-by-trial during the tasks (Graz - TBT and NeuRow - TBT) or during rest (Rest).

This is expected, as task-elicited activity is not only of higher amplitude leading to higher SNR, but it is also a common external driver to both EEG and fMRI signals implicitly enhancing their relationship. Importantly, our work shows that there is still some correlation, albeit weaker, between EEG and BOLD-fMRI in spontaneous fluctuations of brain activity, either during rest or over trials, despite their lower amplitude trial-by-trial fluctuations. Indeed, even though correlations have been reported between the BOLD activity of resting-state networks and concurrent EEG spectral power, they are typically low (about 0.2, as in our study) and of variable consistency \citep{xavier2025}. The slightly better prediction of the trial-by-trial fluctuation conditions compared to the resting-state condition suggests the execution of a task may lead to a stronger coupling between the two modalities even when the task-evoked activities are removed. These results not only provide strong evidence that we are indeed capturing the relationship between the two modalities (without any external drivers), but they are also a testament to the power of our models.

\subsection{Model interpretability}
The MUC patterns (shown in Fig.~\ref{fig:massive_univariate} \& Fig.~\ref{fig:massive_univariate_ext}) were generally consistent with the expected SMR. We found negative correlations of fMRI motor brain activity with EEG power in the alpha and/or beta bands (8 - 30 Hz), consistently with the desynchronization in these frequencies \citep{Pfurtscheller1999}. Interestingly, two sub-bands were clearly evident in our case, one in the lower alpha band, $\sim$8-9 Hz, and another one in the beta band, $\sim$15-20 Hz. These findings may partly explain why broader frequency bands may not elicit good predictors. Also as expected, although this correlation pattern was spread throughout the brain, it was nevertheless strongest in channels C3-C4, near the somatomotor cortex, and also in the parietal channels CP1-CP2 and P3-P4-Pz, as has been previously observed in MI tasks \citep{Batista2024}. Moreover, the hemodynamic lags that produced the strongest correlations were centered around the canonical value of about 6 seconds \citep{Logothetis2001} but varied considerably across frequencies. Altogether, although such univariate computation does not account for interactions between EEG regressors, it nevertheless informed us that, despite a predominance of negative correlations in the alpha/beta ranges around central electrode locations, and for canonical haemodynamic delays, a more complex distribution across frequencies, channels and delays clearly sets the motivation for the features included in our models. 

In fact, by analyzing the learned model coefficients (Fig.~\ref{fig:model_coefficients} \& Fig.~\ref{fig:distributed_lag_ext}), we found a large degree of agreement with the MUC patterns. Interestingly, the learned coefficients show distinctively sparser patterns, suggesting that the SGL penalty was effective in learning a smooth but sparsified approximation of the MUC, with redundant and uninformative predictors attenuated. This effect was especially pronounced spatially, with the channels nearest the somatomotor cortex (Cz-C3-C4) and parietal regions (CP1-CP2 and Pz-P3-P4) exhibiting the strongest negative correlations across all conditions in both $\sim$8-10 Hz and $\sim$15-25 Hz frequency ranges, which suggests our grouping of covariates by EEG channel was effective in promoting spatial sparsity.

\subsection{Limitations and future work}
Limitations of this work arise from certain modelling assumptions. The assumptions of normally distributed and independent errors are violated if the underlying generative process of the BOLD time series is non-Gaussian. Although linear regression is generally robust to most scenarios that violate these assumptions, the independence of errors is directly violated by autocorrelations in BOLD time series. Though non-stationarity null hypotheses were rejected for all preprocessed BOLD time series in our dataset, recent works suggest that frequently-used fMRI preprocessing packages may suboptimally prewhiten fMRI time series \citep{Olszowy2019}, which could admit more false positives during surrogate data testing. To preempt these concerns, we have carefully designed our surrogate data test to remove undesirable inclusion criteria from our null hypothesis. These preemptive measures include using the Augmented Dicky-Fuller test to remove potential false positives due to non-stationarity of the BOLD time series and also the choice of IAAFT surrogates to circumvent false positives arising from non-Gaussianity of the underlying generative process of BOLD time series. Despite these limitations, it should be noted that most related works do not even employ surrogate data and null models to assess statistical significance of model prediction, and therefore do not address any of these issues. More critically, the fact that most studies rely on training and test data retrieved from the same acquisition session for each subject raises further concerns regarding data leakage and the effects of autocorrelations. 

While the choice of a linear model improves interpretability and reliability of statistical inferences made about the implicated physiological time series, nonlinear models could potentially achieve better prediction performance. We have shown that a simple adaptation of the well-established EEGNet model \citep{lawhern2018} to the EEG-to-fMRI regression problem was able to achieve comparable performance to SGL in a more parameter- and sample-efficient manner, using the raw data directly rather than spectral power features \cite{Stabile2025}. Further improvements will certainly be possible by using more sophisticated deep learning models. Given the scarcity of large EEG-fMRI datasets, mostly due to the complexity of the underlying data acquisitions, this will require the combination of multiple datasets across different studies and sites. Another assumption of the linear regression is the stationarity of the EEG-to-fMRI model. Critically, we have recently shown that EEG-fMRI couplings may vary over time scales of seconds to minutes, with rapid switching between distinct correlation states \citep{sotablumenfeld,mendes2025}, warranting the exploration of prediction approaches that are able to account for these.

\subsection{Conclusion}
Our work provides evidence of the ability to predict fMRI motor brain activity from EEG recordings alone across different days, with statistical significance in individual subjects. These results present an important contribution to the literature, highlighting the potential leakage effects of within-session train-test paradigms, while showcasing the ability to predict even spontaneous fluctuations in brain activity across sessions. Overall, these findings hold great promise for translation to EEG neurofeedback applications.

\section{Acknowledgments}
Funding by grants LARSyS (FCT, DOI: 10.54499/LA/P/0083/2020, 10.54499/UIDP/50009/2020), MIGN2Treat (FCT, PTDC/EMD-EMD/29675/2017 and LISBOA-01-0145-FEDER-029675), NeurAugVR (FCT, PTDC/CCI-COM/31485/2017), NOISyS (FCT, 10.54499/2022.02283.PTDC), PRR Center for Responsible AI (C645008882-00000055), and FCT grant SFRH/BD/151128/2021.

\printbibliography

\appendix
\section{Appendix}

\subsection{Event-related Desynchronization}
\begin{figure}[htbp]
    \centering
    \begin{overpic}[scale=0.4]{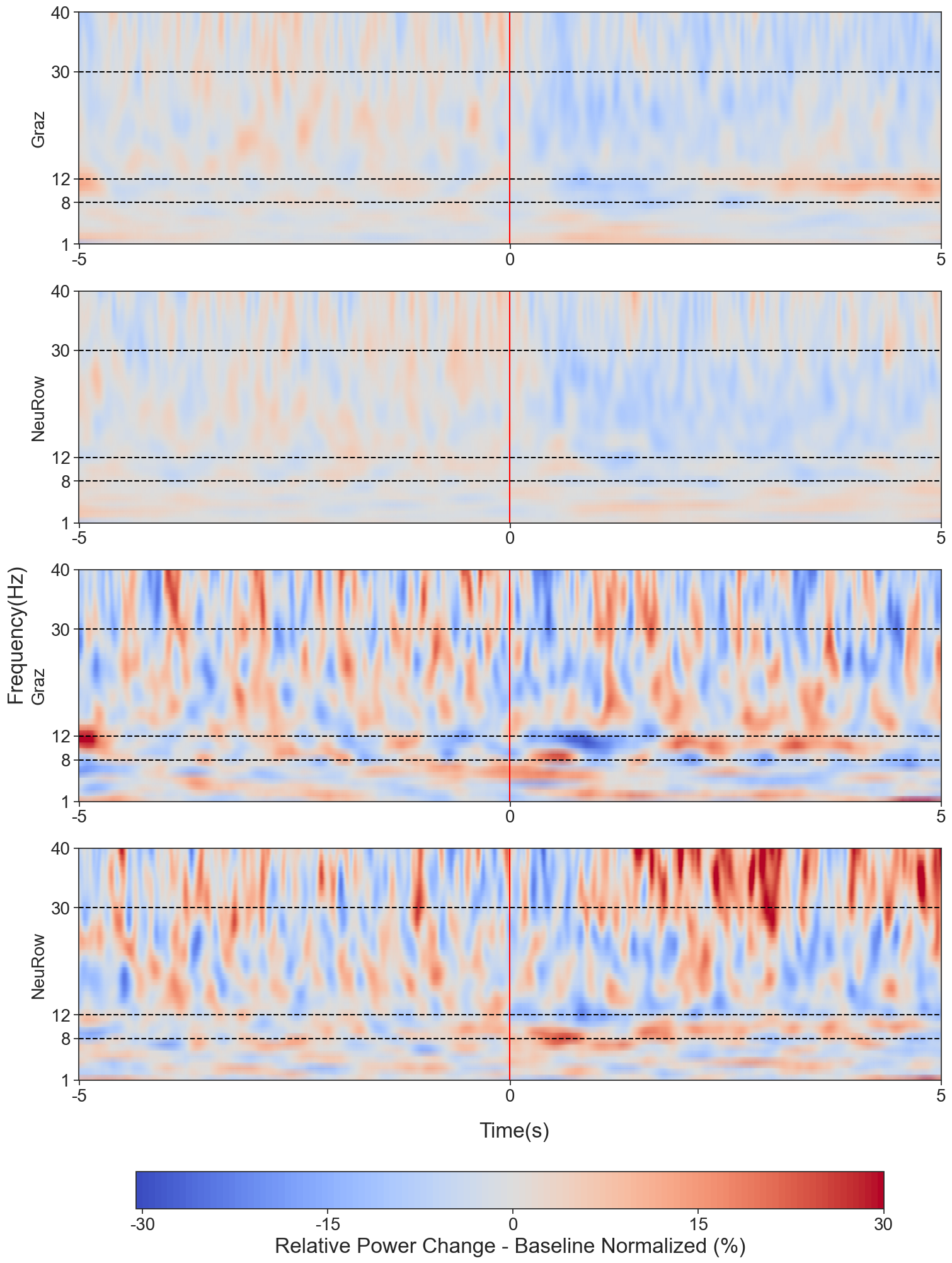}
        \put(1,99){\textbf{A}} 
        \put(1,56){\textbf{B}}
    \end{overpic}    \caption{\textit{Event-Related Desynchronization (ERD).}: (A) Time-frequency plots of time-varying spectral power, baseline normalized, averaged across all subjects. Data are first epoched (-5 s to 5 s around stimulus onset), relative power is estimated within each trial, all trials are averaged, then the result is normalized by a baseline average (-5 s to -0.004s). ERD is clearly demarcated by the blue region just post stimulus. (B) Similar to (A), but for a single representative subject.}
    \label{fig:erd}
\end{figure}

\newpage

\subsection{Hyperparameter Optimization}
\begin{figure}[htbp]
    \centering
    \includegraphics[scale=0.275]{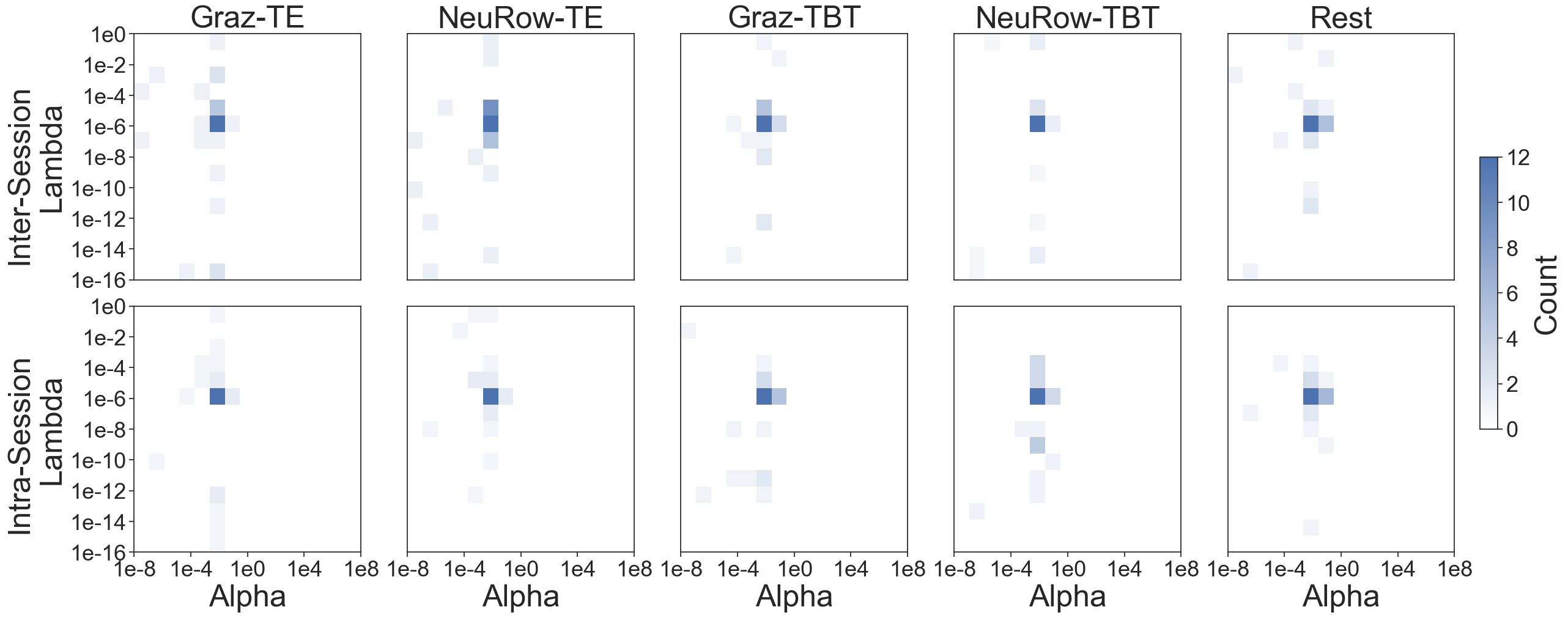}
    \caption{\textit{Frequency of Selected Hyperparameters.}: 2D histograms with darker shade representing higher frequency of selected hyper parameters ($\lambda$ on y-axis, $\alpha$ on x-axis) across subjects and sessions. Both intra- and inter-session models are represented.}
    \label{fig:hyperparams}
\end{figure}

\newpage

\subsection{Statistical Significance of Coefficients}

\begin{figure}[htbp]
    \centering
    \begin{overpic}[scale=0.275]{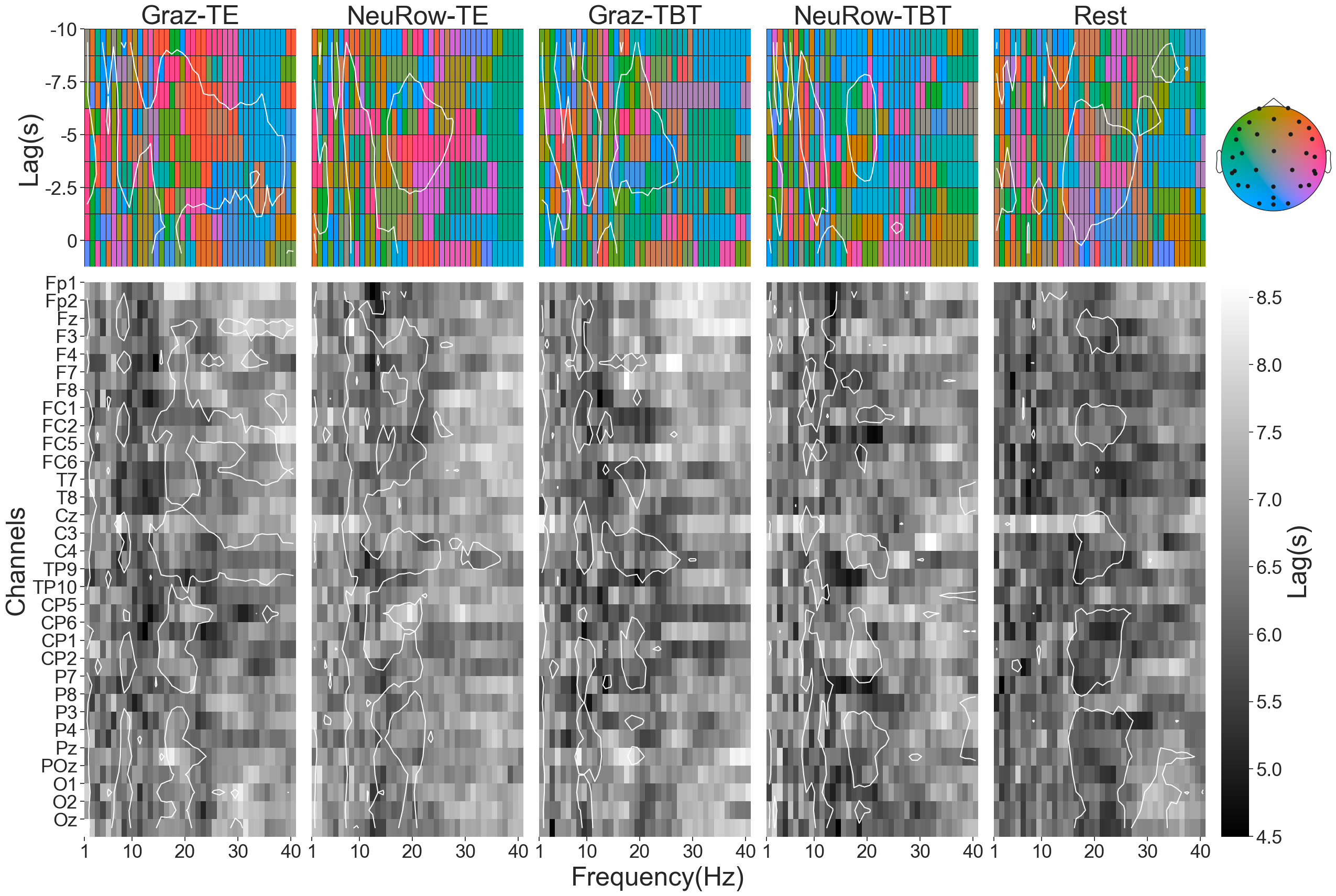}
        \put(1,65){\textbf{A}} 
        \put(1,45){\textbf{B}}
    \end{overpic}
    \caption{\textit{Optimal Channels and Lags for MUC Analyses.}: (A) Most informative channels (mode of channels selected from maximum absolute correlation) visualized across the scalp for specific lags and frequencies. (B) Lags associated with maximum absolute correlation for each channel and frequency.}
    \label{fig:massive_univariate_ext}
\end{figure}

\begin{figure}[htbp]
    \centering
    \begin{overpic}[scale=0.275]{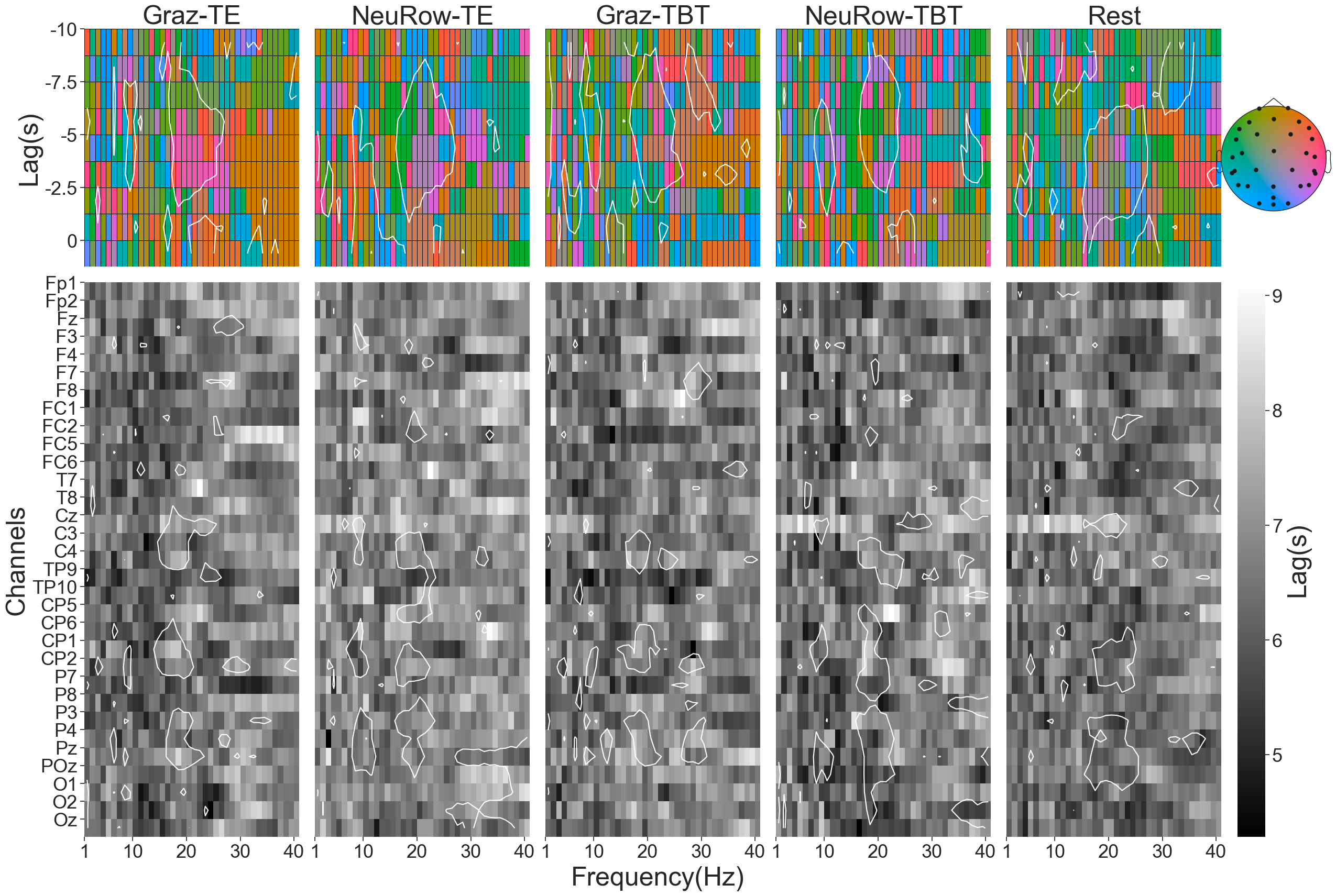}
        \put(1,65){\textbf{A}} 
        \put(1,45){\textbf{B}}
    \end{overpic}
    \caption{\textit{Optimal Channels and Lags for SGL Coefficients.}: (A) Most informative channels (mode of channels selected from maximum absolute coefficient) visualized across the scalp for specific lags and frequencies. (B) Lags associated with maximum absolute coefficient for each channel and frequency.}
    \label{fig:distributed_lag_ext}
\end{figure}

\end{document}